\documentclass[10pt,twocolumn,letterpaper]{article}

\usepackage{wacv}              

\usepackage{graphicx}
\usepackage{amsmath}
\usepackage{amssymb}
\usepackage{booktabs}

\usepackage[pagebackref,breaklinks,colorlinks]{hyperref}

\usepackage[capitalize]{cleveref}
\crefname{section}{Sec.}{Secs.}
\Crefname{section}{Section}{Sections}
\Crefname{table}{Table}{Tables}
\crefname{table}{Tab.}{Tabs.}

\def\wacvPaperID{871} 
\def\confName{WACV}
\def\confYear{2025}

\usepackage{multirow}
\usepackage{algorithm}
\usepackage{algpseudocode}

\begin{document}

\title{Identity Curvature Laplace Approximation for Improved Out-of-Distribution Detection}

\author{Maksim Zhdanov\\
AIRI\thanks{Work done while at T-Bank}, NUST MISIS\thanks{National University of Science and Technology MISIS}\\
Moscow, Russia\\
{\tt\small zhdanov@airi.net}
\and
Stanislav Dereka\\
T-Bank, MIPT\thanks{Moscow Institute of Physics and Technology}\\
Moscow, Russia\\
{\tt\small s.dereka@tbank.ru}
\and
Sergey Kolesnikov\\
T-Bank\textsuperscript{\text{*}}\\
Moscow, Russia\\
{\tt\small scitator@gmail.com}
}

\maketitle

\begin{abstract}

Uncertainty estimation is crucial in safety-critical applications, where robust out-of-distribution (OOD) detection is essential. Traditional Bayesian methods, though effective, are often hindered by high computational demands. As an alternative, Laplace approximation offers a more practical and efficient approach to uncertainty estimation. In this paper, we introduce the Identity Curvature Laplace Approximation (ICLA), a novel method that challenges the conventional posterior covariance formulation by using identity curvature and optimizing prior precision. This innovative design significantly enhances OOD detection performance on well-known datasets such as CIFAR-10, CIFAR-100, and ImageNet, while maintaining calibration scores. We attribute this improvement to the alignment issues between typical feature embeddings and curvature as measured by the Fisher information matrix. Our findings are further supported by demonstrating that incorporating Fisher penalty or sharpness-aware minimization techniques can greatly enhance the uncertainty estimation capabilities of standard Laplace approximation.

\end{abstract}

\section{Introduction}


Deep neural networks have become essential tools across various machine learning domains, including computer vision \cite{dosovitskiy2020image}, natural language processing \cite{vaswani2017attention}, and time-series forecasting \cite{zhang2023effectively}. Despite their wide application, these networks often exhibit overconfidence in their predictions \cite{nguyen2015deep}, posing challenges in safety-critical environments. Uncertainty estimation, which evaluates a neural network's confidence in its predictions, addresses this issue by providing a measure of trustworthiness. A key application is out-of-distribution (OOD) detection, which identifies samples that significantly differ from the training data \cite{hendrycks2018deep}, typically resulting in unreliable predictions due to high overconfidence \cite{zhang2023openood}.

\begin{figure}[t!]
\includegraphics[width=\linewidth]{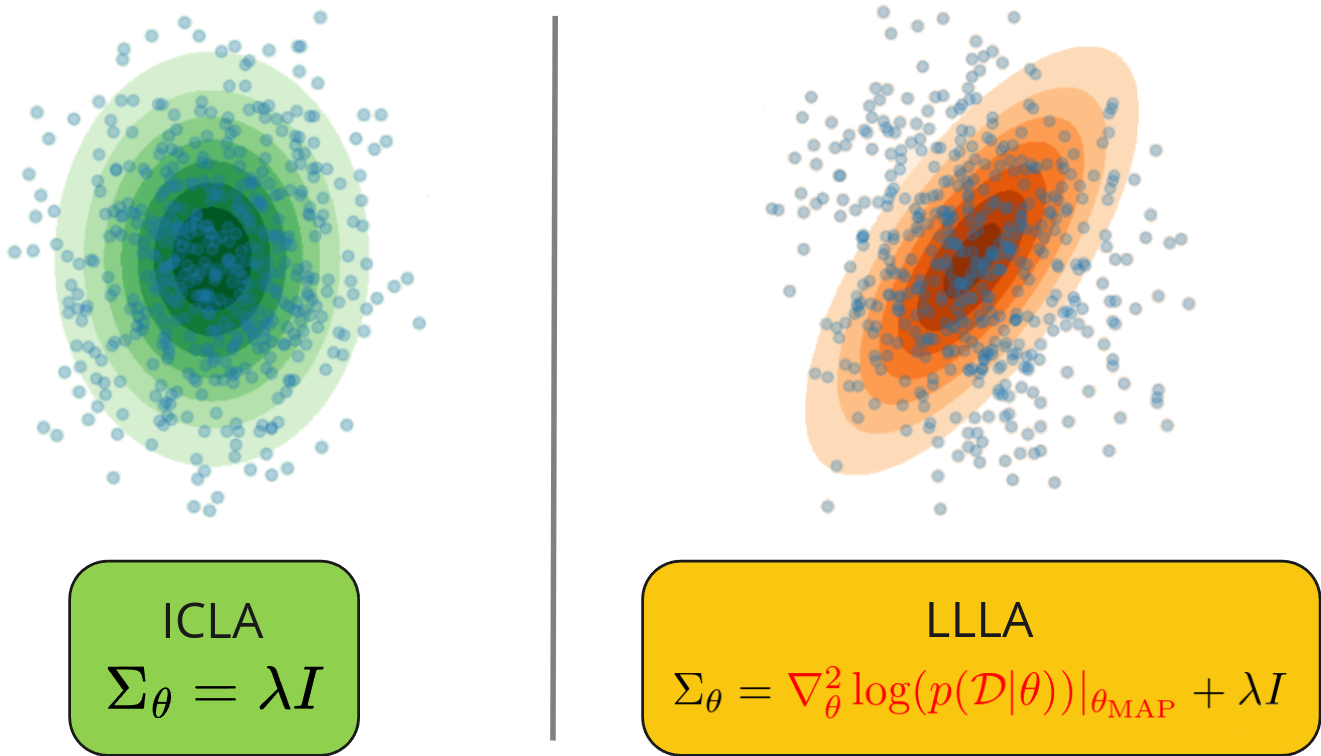}
\caption{Identity Curvature Laplace Approximation (ICLA), without Hessian computation, produces smoother uncertainty landscapes than standard last-layer Laplace approximation (LLLA), and leads to improved OOD detection performance in real-world scenarios.}
\label{fig:toy}
\end{figure}

\begin{figure*}[t]
\centering
\includegraphics[width=0.8\textwidth]{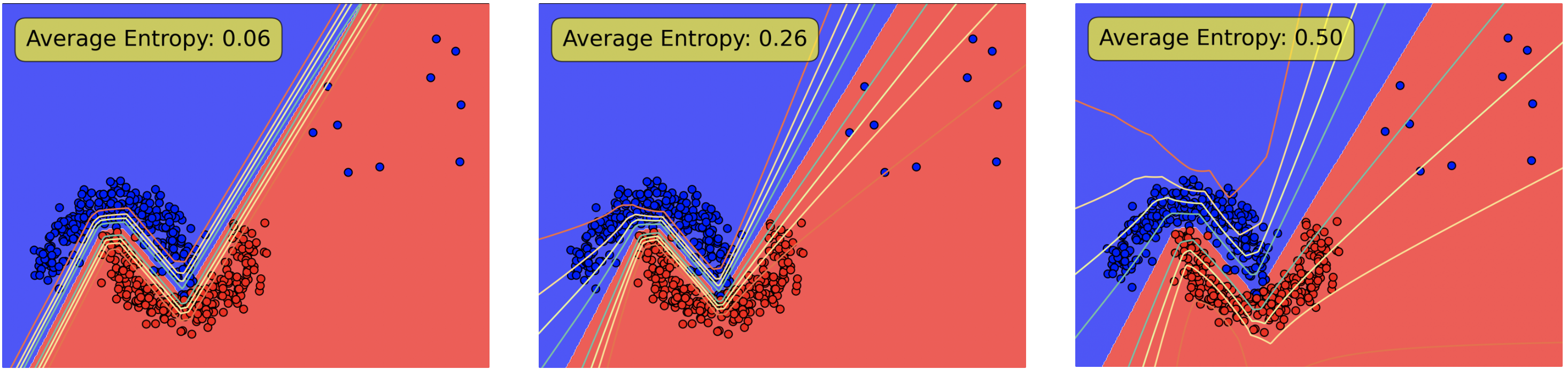} 
\caption{Uncertainty estimation on a toy binary classification dataset with test outliers for MAP network (\textit{left}), Standard LLLA (\textit{middle}), ICLA (\textit{right}). For each setup, the average predictive uncertainty for outlier samples is displayed in the top-left corner. It can be seen that using identity curvature and prior precision as the source of covariance leads to a wider uncertainty surface while preserving the posterior landscape. Subsequently, it achieves better entropy estimates for outlier data points. See Section \ref{sec:toy}.}
\label{halfmoons}
\end{figure*}


Bayesian methods \cite{neal2012bayesian, hinton1993keeping} have been popular for addressing uncertainty in neural networks \cite{jospin2022hands}. The Laplace approximation \cite{mackay1992practical}, a computationally efficient Bayesian technique, approximates the posterior distribution of model weights with a Gaussian centered at the point of maximum likelihood. Efficient variations of this method \cite{ritter2018scalable, ritter2018online} make Bayesian inference feasible for large networks. For example, the last-layer Laplace approximation \cite{kristiadi2020being} reduces computational complexity by focusing on the final layer's weights.


However, some researchers question the necessity of complex Bayesian methods. Deep ensembles (DE), a straightforward non-Bayesian approach, have shown comparable or superior performance in uncertainty estimation \cite{lakshminarayanan2017simple}, suggesting that simpler methods might be sufficient for practical applications.


In this work, we identify an intrinsic limitation of model curvature in Laplace approximation for uncertainty estimation. We propose the Identity Curvature Laplace Approximation (ICLA\footnote{Source code: \href{https://github.com/maxnygma/icla}{https://github.com/maxnygma/icla}}), which replaces the Hessian with an identity matrix and optimizes prior precision, enhancing OOD detection while maintaining calibration scores. Unlike traditional methods, ICLA eliminates the need to compute the Hessian matrix during inference (see Figure \ref{fig:toy}). We empirically demonstrate that class separability in the embedding space explains ICLA's effectiveness. Furthermore, we show that applying a Fisher trace penalty or sharpness-aware minimization during training smooths the model parameter landscape, closing the performance gap between standard Laplace approximation and ICLA.

The contributions of this work can be summarized as follows:

\begin{itemize} 
    \item We introduce ICLA, a simplified Laplace approximation method that optimizes prior precision without requiring Hessian computations at inference. ICLA outperforms standard Laplace methods in OOD detection on CIFAR-10, CIFAR-100, and ImageNet-200 datasets, while maintaining similar calibration metrics.
    \item We highlight the connection between curvature and class separability in the embedding space, explaining ICLA's improved performance.
    \item We demonstrate that employing a Fisher penalty or sharpness-aware minimization techniques during training enhances the smoothness of the model parameters landscape, thereby improving the performance of traditional Laplace approximation methods.
\end{itemize}

\section{Background}

\begin{figure*}[t]
\centering
\includegraphics[width=0.8\textwidth]{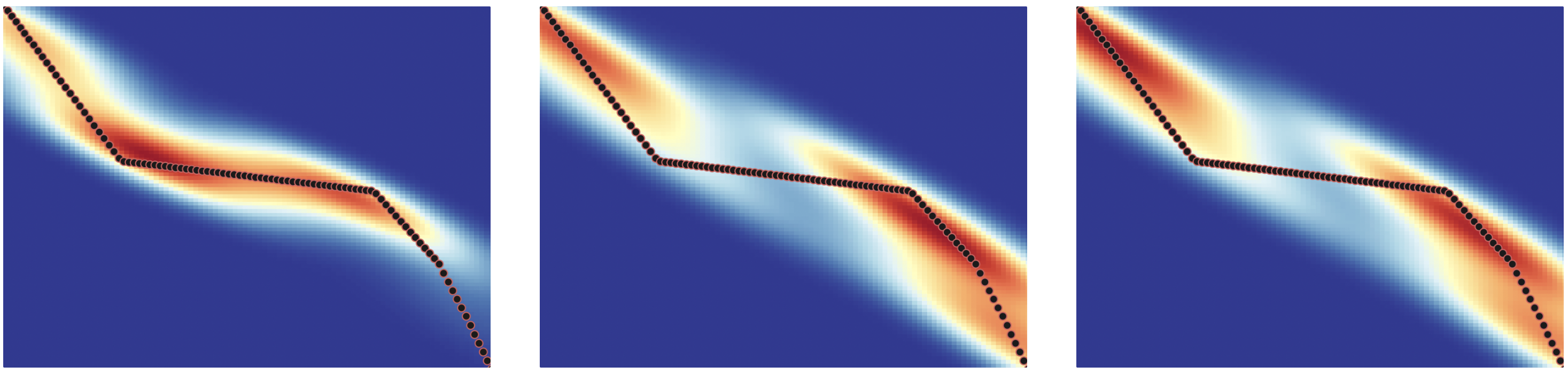} 
\caption{Uncertainty estimation on a toy sinusoidal regression dataset for MAP network (\textit{left}), Standard LLLA (\textit{middle}), ICLA (\textit{right}). The observed property of Laplace approximation and identity curvature generalizes to regression tasks and does not negatively impact uncertainty estimates. See Section \ref{sec:toy}.}
\label{sin}
\end{figure*}

\paragraph{Out-of-Distribution Detection.}

We consider a multi-class classification problem where $\mathcal{D}=\{x_i, y_i\}^N_{i=1}$ represents a dataset consisting of independently and identically distributed (i.i.d.) data points $x_i \in \mathbb{R}^D$ and labels $y \in \{k_1, \dots, k_C\}$, each depicted as a one-hot vector in a $C$-dimensional space.

Given the defined multi-class classification problem, we have a particular set of classes $y$ that we want to model. We refer to these classes and the corresponding data samples as \textit{in-distribution} (ID) data. In an open world, however, other groups of classes do not belong to our training data $\mathcal{D}$. We can say that they come from an \textit{out-of-distribution} (OOD) domain. The task of OOD detection is to measure how well a predictive model can identify whether a data sample at inference phase is from an ID or OOD domain \cite{zhang2023openood}. 

\paragraph{Calibration.} 

A neural network is considered calibrated when its output probabilities align with its real predictive performance. For instance, if for some set of samples $\{x_{i, i+n}\}$ a model predicts the label $y$ with probability $P$, then we expect for a fraction of $P$ of all data points from $\{x_{i, i+n}\}$ to have $y$ as a true label. Common calibration measures are the expected calibration error (ECE) \cite{naeini2015obtaining}, negative log-likelihood (NLL) \cite{guo2017calibration} and Brier score \cite{minderer2021revisiting}. Refer to Appendix \ref{app:calibration} for the definitions.

\paragraph{Bayesian Models.}

We define a neural network as $f_{\theta}: \mathbb{R}^D \rightarrow \mathbb{R}^C$ parameterized by $\theta$, with a likelihood $p(\mathcal{D}|\theta)$, as well as prior $p(\theta)$ and posterior distribution $p(\theta|\mathcal{D})~\propto~p(\mathcal{D}|\theta)p(\theta)$. According to Bayesian principles, the standard training of a neural network is a \textit{maximum a posteriori} (MAP) estimation. This involves finding $\theta_{\text{MAP}}~=~\operatorname*{arg\,max}_{\theta}p(\theta|\mathcal{D})$, which can be further expanded as $\operatorname*{arg\,max}_{\theta}\operatorname{log}(p(\mathcal{D}|\theta))+\operatorname{log}(p(\theta))$. 

In practice, the log-likelihood $\operatorname{log}(p(\mathcal{D}|\theta))$ corresponds to cross-entropy loss for classification tasks, and the prior $\operatorname{log}(p(\theta))$ is typically an isotropic Gaussian distribution $\mathcal{N}(\theta|0,\sigma^2I)$. In neural networks, a Gaussian prior can be achieved by applying weight decay regularization $w(\theta) = \frac{1}{2}\sigma^{-2} \|\theta\|^2$. If the prior is uniform, then training is equivalent to simple \textit{maximum likelihood estimation} (MLE).

Since exact $p(\theta|\mathcal{D})$ posterior estimation is intractable for modern neural networks, many methods were proposed to approximate Bayesian inference, including the Monte Carlo dropout \cite{gal2016dropout} and deep ensembles \cite{wilson2020bayesian, lakshminarayanan2017simple}. Unfortunately, when compared to standard training methods, many of them have a significant computational cost.

\paragraph{Laplace Approximation.}

The simple, yet competitive approximation of the posterior $q(\theta|\mathcal{D})$ can be achieved using \textit{Laplace approximation} \cite{mackay1992practical}. This method computes a second-order expansion of $\operatorname{log}(p(\mathcal{D}|\theta))$ around $\theta_{\text{MAP}}$, resulting in a Gaussian approximation to $p(\theta|\mathcal{D})$. $\operatorname{log}(p(\mathcal{D}|\theta))$ decomposes as

\begin{multline}
\operatorname{log}(p(\mathcal{D}|\theta))\approx \\ \approx \operatorname{log}(p(\mathcal{D}|\theta_{\text{MAP}})) + \frac{1}{2}(\theta-\theta_{\text{MAP}})^T\Sigma_{\theta}(\theta-\theta_{\text{MAP}}),
\end{multline}
\begin{equation}\label{eq:hessian}
    \Sigma_{\theta} \triangleq \nabla^2_{\theta}\operatorname{log}(p(\mathcal{D}|\theta))|_{\theta_{\text{MAP}}} + \lambda I,
\end{equation}
where $\Sigma_{\theta}$ is the Hessian of the log-likelihood with respect to $\theta$ evaluated at $\theta_{\text{MAP}}$ and with prior precision $\lambda$ times identity matrix $I$. In practice, $\lambda$ optimization is achievable using methods such as cross-validation or marginal likelihood \cite{immer2021improving, ritter2018scalable, immer2021scalable}.
Thus, we approximate $p(\theta|\mathcal{D})$ as $\mathcal{N}(\theta|\theta_{\text{MAP}},\Sigma_{\theta}^{-1})$.

\paragraph{Hessian Approximations.}\label{par:hessian}

For modern neural networks, computing the full Hessian is not feasible, since Equation \ref{eq:hessian} scales quadratically with the number of parameters $\theta$. Fortunately, there are multiple approaches for efficient computation of the Hessian in deep neural networks. The most popular methods include the generalized Gauss-Newton (GGN) matrix \cite{schraudolph2002fast} and the Fisher information matrix \cite{amari1998natural}, as well as block-diagonal factorizations such as the Kronecker-factored approximate curvature (K-FAC) \cite{martens2015optimizing} (Refer to Appendix \ref{app:hessian} for the definitions). There is also empirical Fisher (EF), which uses expectations over samples \cite{daxberger2021laplace}. It is important to note that Fisher is the same as GGN when utilizing distributions from the exponential family \cite{martens2020new}, making them both well-justified approximations of the Hessian.

\paragraph{Last-Layer Laplace Approximation.}

Studies have shown that Bayesian methods applied solely to the final layer can rival the performance of those applied across all layers \cite{kristiadi2020being, snoek2015scalable}. In the context of last-layer Laplace approximation (LLLA) \cite{kristiadi2020being}, the neural network $f_{\theta}$ is viewed as $f(x_i)=W\nu(x_i)$, where $W \in \mathbb{R}^{C \times L}$ is the weight matrix of the last layer, and $\nu \in \mathbb{R}^L$ represents the output from the penultimate layer. This results in a covariance $\Sigma_{\theta} \triangleq (\nabla^2_{\theta}\operatorname{log}(p(\mathcal{D}|\theta))|_{W_{\text{MAP}}} + \lambda I) \in \mathbb{R}^{CL \times CL}$.
In case of linearized predictive distribution \cite{immer2021improving} in the last-layer approximation, the output distribution is given by 
\begin{multline}\label{eq:output_distribution}
    p(f(x_i)|x_i,\mathcal{D}) = \\ = \mathcal{N}(f(x_i)|W_{\text{MAP}}\nu(x_i), J(x_i)^T\Sigma^{-1} J(x_i)),
\end{multline}
where $J$ is the Jacobian at input $x_i$. Note that this approach does not require the Monte Carlo integration, which is often inaccurate in case of Laplace approximation \cite{immer2021improving}.

\section{Method}\label{sec:method}

We put forward the hypothesis that, in widely adopted neural network models, \textit{curvature may act as a bottleneck in terms of uncertainty estimation capabilities}. In particular, model curvature may conflict with class covariance typically present in classification tasks. Based on our observations, we further propose a simplified approach without these limitations.

\paragraph{Analysis approach.}
To confirm our hypothesis, we need a way to measure \textit{model curvature} and \textit{class covariance}. We analyze model curvature with the empirical Fisher information matrix, which is given by 

\begin{equation}\label{eq:emp_fisher}
    F \triangleq \sum^{N}_{n=1} \nabla_{\theta} \operatorname{log}(p_{\theta}(y_{n}|x_{n})) \nabla_{\theta} \operatorname{log}(p_{\theta}(y_{n}|x_{n}))^T.
\end{equation}

We employ the lightweight diagonal representation \cite{denker1990transforming} of the Fisher. This particular form of approximation is chosen for computational accessibility.

To evaluate class covariance we propose a mean class-wise cosine similarity (MCCS) measure: 

\begin{equation}\label{eq:mccs}
\text{MCCS} = \frac{1}{\widetilde{N}} \sum_{\substack{n>m\\n,m \in \overline{1, C}}} \sum_{\substack{i>j\\i,j \in \overline{1, N_C}}} \frac{\nu_{ni} \cdot \nu_{mj}}{\left\| \nu_{ni}\right\| _{2}\left\| \nu_{mj}\right\| _{2}},
\end{equation}
where $\nu_{ni}$ represents the feature of $i$-th element of class $n$, $N_C$ is the number of elements in class, and \mbox{$\widetilde{N}=CN_C(C-1)(N_C-1)/4$}.

Our subsequent analysis in Section \ref{sec:data} shows that the Fisher information matrix usually has a long-tailed spectral distribution, where a few eigenvalues are incredibly large \cite{karakida2019universal}. Moreover, we show that LLLA performance degrades on highly separable datasets. To connect the line of thought between class separability, curvature, and long-tailed Fisher eigenvalues distribution, we hypothesize that \textit{penalizing Fisher long-tailness can improve LLLA performance}. Thus, we apply the Fisher penalty in the model training:

\begin{equation}\label{eq:fp}
    \text{FP} = \alpha \left\| \frac{1}{M} \sum_{i=1}^M \nabla_{\theta} \operatorname{log}(p_{\theta}(y_{i}|x_{i})) \right\|_2,
\end{equation}
where $M$ is the batch size of $(x_i, y_i) \sim \mathcal{D}$ and $\alpha$ is a hyperparameter controlling the regularization strength. In essence, this penalty acts as model curvature regularization, which leads to a \textit{flat minima} by explicitly lowering Fisher matrix long-tailness \cite{jastrzebski2021catastrophic, hochreiter1997flat, keskar2016large, dziugaite2017computing, norton2021diametrical}. As an alternative to the Fisher penalty, we also apply Adaptive Sharpness-Aware Minimization (ASAM) \cite{kwon2021asam}, a widely adopted optimization technique for achieving a flatter curvature landscape. Our analysis shows that by applying these techniques (i.e. reducing Fisher long-tailness), LLLA performance can be improved on highly separable datasets, which confirms our intuition on model curvature and class separability.

\paragraph{Identity Curvature Laplace Approximation.}
Putting all the empirical observations together we come up with a simplification of LLLA, devoid of the above problems. Consider the covariance matrix $\Sigma_{\theta}$ from Equation \ref{eq:hessian}, which is the sum of the Hessian and the product of prior precision with the identity matrix.

We propose constructing a posterior approximation based exclusively on identity curvature and optimized prior precision:
\begin{equation}\label{eq:icla}
    \Sigma_{\theta}= \lambda I.
\end{equation}

\begin{figure*}[t]
\centering
\includegraphics[width=0.9\textwidth]{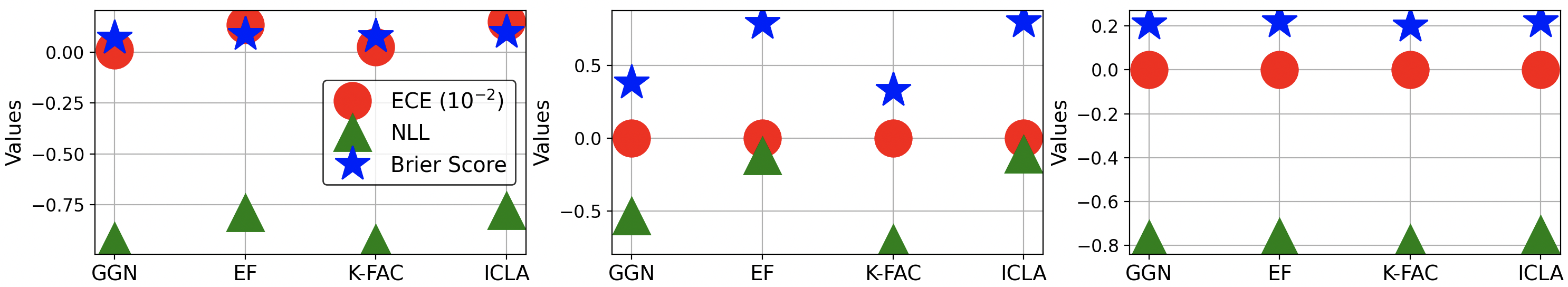} 
\caption{Calibration scores. Expected calibration error (ECE), negative log-likelihood (NLL), and Brier score for CIFAR-10 (\textit{left}), CIFAR-100 (\textit{middle}) and ImageNet-200 (\textit{right}) for different Laplace approximations. Identity curvature does not detrimentally affect model calibration. See Section \ref{sec:acc_and_calib} for details.}
\label{calibration}
\end{figure*}

In this study, we name the proposed method \textit{Identity Curvature Laplace Approximation} (\textbf{ICLA}). This approach doesn't utilize Hessian approximations and is not subject to the long-tailness problem by design. In subsequent experiments, we empirically show that ICLA yields a broader uncertainty landscape and increased entropy for out-of-distribution samples, which facilitates OOD detection. Based on our observations in Section \ref{sec:data}, \textit{ICLA provides the highest performance boosts on highly separable datasets}.

Our Laplace approximation implementation is based on \cite{daxberger2021laplace}, and the algorithm for ICLA training is available in Algorithm \ref{lst:icla}. The function \verb|fit_hessian_approximation| computes diagonal Fisher Hessian approximation for each batch in the validation dataset and returns the mean Hessian value of all batches. The function \verb|marginal_likelihood| performs prior precision optimization (Equation \ref{eq:hessian}) with marginal likelihood method \cite{daxberger2021laplace}, the algorithm is presented in Appendix \ref{app:algs}. Moreover, step 2 of the Algorithm \ref{lst:icla} can be bypassed, i.e. the Hessian matrix is not fitted and initialized with zeros. We call this approach ICLA\textsubscript{zero}. For the full details of implementation, refer to the released ICLA code implementation\footnote{Anonymous submission. Check the supplementary materials}. We also report ICLA computational performance in Appendix \ref{app:compute}.

\begin{algorithm}[t!]
\caption{ICLA training}\label{lst:icla}
\begin{algorithmic}[1]

\Require{Dataset $\mathcal{D}=\{x_i, y_i\}^N_{i=1}$, neural network f, functions \verb|fit_hessian_approximation| and \verb|marginal_likelihood| from \cite{daxberger2021laplace}.}
\Ensure{Prior precision $\lambda$ for Equation \ref{eq:icla}.}
\Statex
\State Initialize $\lambda \gets 1$, Hessian $\mathcal{H} \gets \bf 0$
\State $\mathcal{H} \gets$ \verb|fit_hessian_approximation|$(f, \mathcal{D})$
\State $\lambda \gets$ \verb|marginal_likelihood|$(f, \mathcal{H})$
\State $\mathcal{H} \gets \bf 0$
\State \Return $\lambda$, $\mathcal{H}$

\end{algorithmic}
\end{algorithm}

\section{Experiments}

\begin{table*}[]
\renewcommand{\arraystretch}{1.2} 
\setlength{\tabcolsep}{5pt} 
\centering
\caption{OOD detection results for different Laplace approximations. 
All approximations are tested on multiple OOD sources. The mean and standard deviations' AUROC values are reported. 
The best results among different methods are highlighted.}
\vskip 0.15in
\begin{tabular}{cl|cc|cl|ll}
\hline
\multicolumn{2}{c|}{\multirow{2}{*}{Method}} & \multicolumn{2}{c|}{CIFAR-10} & \multicolumn{2}{c|}{CIFAR-100} & \multicolumn{2}{c}{ImageNet-200} \\
\multicolumn{2}{c|}{} & Near OOD & Far OOD & Near OOD & \multicolumn{1}{c|}{Far OOD} & \multicolumn{1}{c}{Near OOD} & \multicolumn{1}{c}{Far OOD} \\ \hline
\multicolumn{2}{c|}{LLLA (GGN)} & 88.94 ± 0.30 & 91.54 ± 0.36 & \multicolumn{1}{l}{81.44 ± 0.09} & 80.00 ± 0.45 & 81.84 ± 0.10 & 89.39 ± 0.15 \\
\multicolumn{2}{c|}{LLLA (EF)} & \multicolumn{1}{l}{89.54 ± 0.36} & \multicolumn{1}{l|}{92.09 ± 0.30} & \multicolumn{1}{l}{\textbf{81.66 ± 0.10}} & 80.17 ± 0.83 & 81.86 ± 0.11 & \textbf{89.45 ± 0.14} \\
\multicolumn{2}{c|}{LLLA (K-FAC)} & \multicolumn{1}{l}{88.02 ± 0.28} & \multicolumn{1}{l|}{90.73 ± 0.52} & 80.30 ± 0.13 & 77.77 ± 0.53 & 81.73 ± 0.10 & 88.94 ± 0.14 \\ \hline
\multicolumn{2}{c|}{\textbf{ICLA}} & \textbf{90.01 ± 0.21} & \textbf{92.50 ± 0.38} & 81.45 ± 0.10 & \multicolumn{1}{c|}{\textbf{80.79 ± 0.46}} & \textbf{81.88 ± 0.10} & \textbf{89.45 ± 0.14} \\ \hline
\end{tabular}
\vskip -0.1in
\label{llla-table}
\end{table*}

\begin{table*}[]
\renewcommand{\arraystretch}{1.2} 
\setlength{\tabcolsep}{5pt} 
\centering
\caption{OOD detection results against non-Bayesian methods. All methods are trained on the ID dataset and tested on multiple OOD sources. The mean and standard deviations' AUROC values are reported. The best results among different methods are highlighted and the 2nd best are underlined.}
\vskip 0.15in
\begin{tabular}{cl|cc|cc|ll}
\hline
\multicolumn{2}{c|}{\multirow{2}{*}{Method}} & \multicolumn{2}{c|}{CIFAR-10} & \multicolumn{2}{c|}{CIFAR-100} & \multicolumn{2}{c}{ImageNet-200} \\
\multicolumn{2}{c|}{} & Near OOD & Far OOD & Near OOD & Far OOD & \multicolumn{1}{c}{Near OOD} & \multicolumn{1}{c}{Far OOD} \\ \hline
\multicolumn{2}{c|}{MSP} & 88.03 ± 0.25 & 90.73 ± 0.43 & 80.27 ± 0.11 & 77.76 ± 0.44 & \textbf{83.34 ± 0.06} & 90.13 ± 0.09 \\
\multicolumn{2}{c|}{ODIN} & 82.87 ± 1.85 & 87.96 ± 0.61 & 79.90 ± 0.11 & 79.28 ± 0.21 & 80.27 ± 0.08 & 91.71 ± 0.19\\
\multicolumn{2}{c|}{ReAct} & 87.11 ± 0.61 & 90.42 ± 1.41 & 80.77 ± 0.05 & 80.39 ± 0.49 & 81.87 ± 0.98 & \underline{92.31 ± 0.56}\\
\multicolumn{2}{c|}{VIM} & 88.68 ± 0.28 & \textbf{93.48 ± 0.24} & 74.98 ± 0.13 & \underline{81.70 ± 0.62} & 78.68 ± 0.24 & 91.26 ± 0.19 \\
\multicolumn{2}{c|}{SHE} & 81.54 ± 0.51 & 85.32 ± 1.43 & 78.95 ± 0.18 & 76.92 ± 1.16 & 80.18 ± 0.25 & 89.81 ± 0.61 \\
 \multicolumn{2}{c|}{ASH}& 75.27 ± 1.88& 78.49 ± 2.31& 78.20 ± 2.21& 80.58 ± 2.56& 79.38 ± 1.93&\textbf{92.74 ± 1.91}\\ \hline
\multicolumn{2}{c|}{\textbf{ICLA}} & \underline{90.01 ± 0.21} & 92.50 ± 0.38 & \textbf{81.45 ± 0.10} & 80.79 ± 0.46 & \underline{81.88 ± 0.10} & 89.45 ± 0.14 \\
\multicolumn{2}{c|}{\textbf{ICLA}\textsubscript{zero}} & \textbf{90.56 ± 0.23} & \underline{93.18 ± 0.45} & \underline{81.38 ± 0.28} & \textbf{82.49 ± 0.60} & 80.70 ± 0.10 & 89.82 ± 0.11 \\ \hline
\end{tabular}
\vskip -0.1in
\label{non-bayes-table}
\end{table*}


\subsection{Synthetic Examples}

\paragraph{Half-Moons.}\label{sec:toy}

To visually examine the changes ICLA brings to uncertainty estimates, we use a half-moons binary classification dataset with an additional 10 test outliers. In this experiment, we use a simple 2-layer MLP with ReLU activations and 20 hidden units. We compare the uncertainty landscapes produced by 3 methods: MAP, LLLA, and ICLA. The results are presented in Figure \ref{halfmoons}. We demonstrate that ICLA preserves the general structure of the predictive distribution and gives more strict and precise uncertainty estimates for outliers. In addition, one can see that ICLA's uncertainty surface leads to higher outlier entropy scores, which is crucial in OOD detection task.

\paragraph{Sinusoidal Regression.}

It is essential to make sure that our observations are not limited to toy classification settings and are capable of generalizing to synthetic regression. We visually demonstrate this in Figure \ref{sin} using a sinusoidal regression dataset. For this experiment, we use the same network as in the previous setup and compare the same set of methods: MAP, LLLA and ICLA.
We show that ICLA produces uncertainty estimates similar to LLLA, which involves Hessian computation.

\subsection{OOD Detection}\label{sec:ood}

For OOD detection experiments, we employ the well-established OpenOOD v1.5 \cite{zhang2023openood} benchmark, which features the most state-of-the-art OOD detection methods. It uses CIFAR-10, CIFAR-100, and ImageNet-200 (a subset of ImageNet-1K) \cite{deng2009imagenet} as in-distribution datasets and evaluates methods against carefully chosen OOD datasets. It has “Near OOD” and “Far OOD” parts, where the former contains OOD datasets semantically more similar to ID and the latter contains OOD datasets of completely different categories from ID data. For all of our datasets, we use ResNet-18 \cite{he2016deep} trained with the SGD optimizer with a momentum of 0.9. The initial learning rate is set to 0.1 and then reduced to $10^{-6}$ with the Cosine Annealing scheduler \cite{loshchilov2016sgdr}. Training lasts 100 epochs on CIFAR-10 and CIFAR-100 and 90 epochs on ImageNet-200. Additionally, we perform a multi-seed evaluation with 3 random seeds and report the mean and standard deviation values for each experiment to obtain a fair comparison. The predictive distribution for OOD samples is obtained in the same manner as for estimation of ID uncertainty.

\paragraph{Laplace Approximations.}

We compare standard LLLA with GGN, EF, and K-FAC against ICLA in terms of OOD detection AUROC \cite{zhang2023openood}. We tune prior precision for all of our methods via the marginal likelihood method \cite{immer2021scalable}. The results are presented in Table \ref{llla-table}. It can be seen from the results that ICLA achieves better performance on OOD detection problems than other Laplace approximation variations in 5 of 6 experiments.

\paragraph{Non-Bayesian Approaches.} 

In addition to Laplace approximations, we compare ICLA with popular post-hoc non-Bayesian OOD detection methods from OpenOOD, including Maximum Softmax Probability (MSP) \cite{hendrycks2016baseline}, ODIN \cite{liang2017enhancing}, ReAct \cite{sun2021react}, VIM \cite{wang2022vim} SHE \cite{zhang2022out}, and ASH \cite{djurisic2022extremely}. The comparison is shown in Table \ref{non-bayes-table}. ICLA outperforms the mentioned methods on “Near OOD” CIFAR-10 and CIFAR-100 and achieves competitive performance on other setups.

\subsection{Accuracy and Calibration}\label{sec:acc_and_calib}

Employing identity curvature in Laplace approximation does not affect the in-distribution classification accuracy, as the method maintains an optimal solution around $\theta_{\text{MAP}}$ \cite{kristiadi2020being}. Most importantly, it does not significantly change in-distribution probabilities, impacting calibration scores (see Figure \ref{calibration}). To compare calibrations, we use the negative log-likelihood (NLL), expected calibration error (ECE) and Brier score. ICLA maintains calibration metrics relatively similar to more computationally complex versions of Laplace approximations. 

\section{Analysis}\label{sec:data}

\begin{figure}[t]
\centering
\includegraphics[width=0.9\columnwidth]{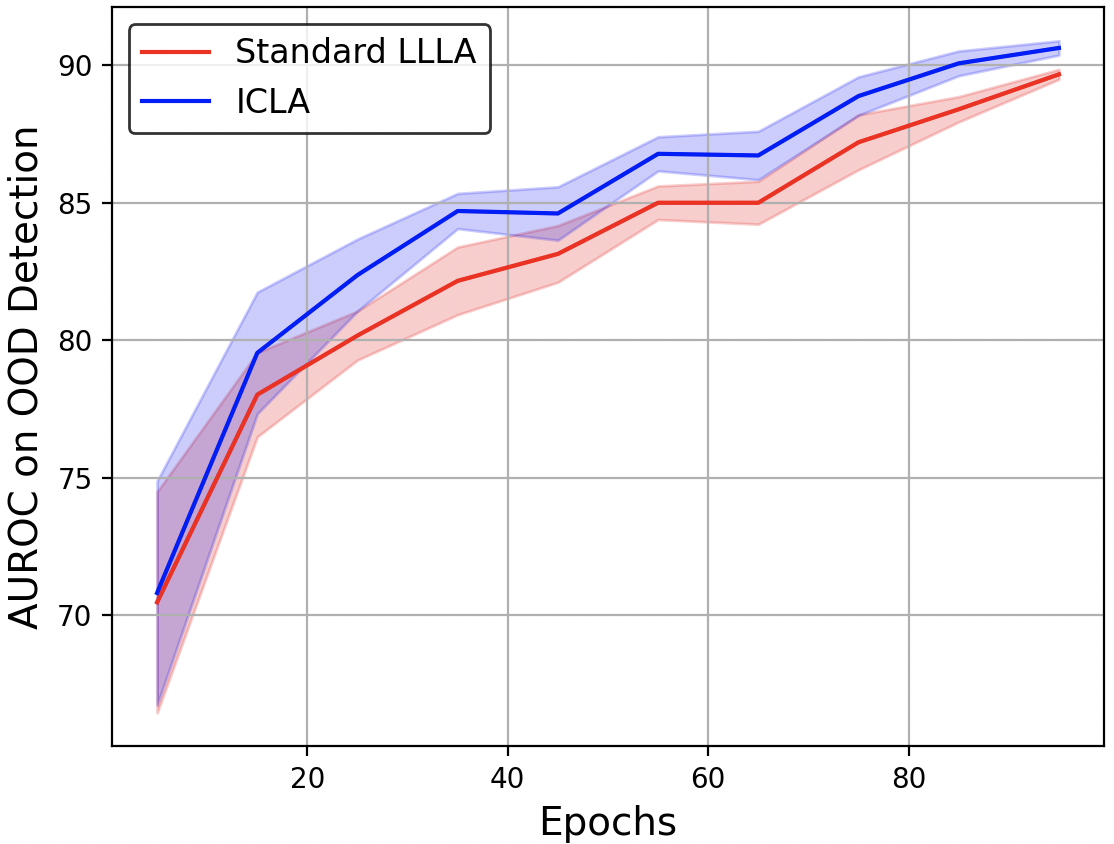} 
\caption{Epoch-wise comparison between standard LLLA and ICLA for OOD detection. Based on this experiment, we can conclude that the observed phenomenon is unrelated to the model training stage. See Section \ref{sec:training_stage} for details.}
\label{epochwise}
\end{figure}

\begin{figure}[t]
\centering
\includegraphics[width=\columnwidth]{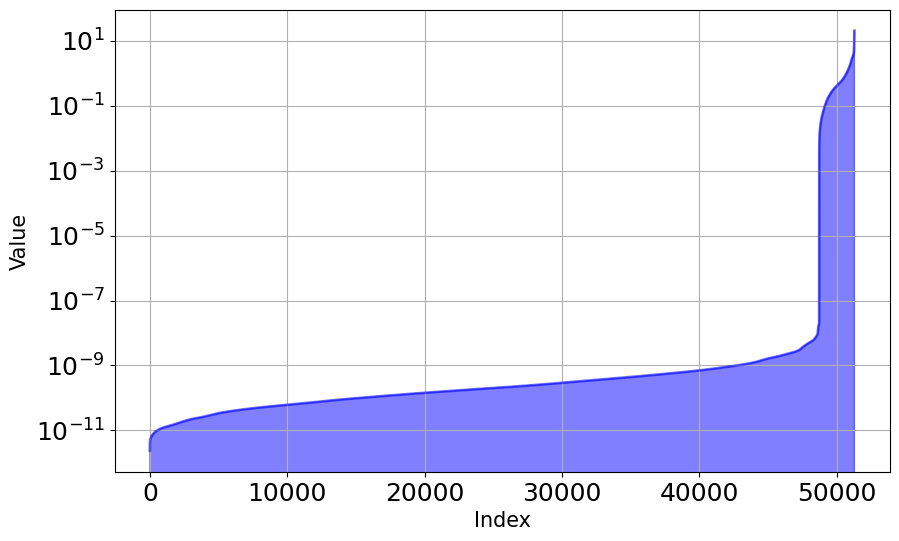} 
\caption{Diagonal empirical Fisher eigenvalues. Only a small percentage of eigenvalues represent large values, leading us to conclude that the spectral distribution has a long tail. Index stands for the i-th eigenvalue. See Section \ref{sec:fisher_eig} for details.}
\label{spectral}
\end{figure}

To understand the roots of the efficacy of ICLA, we apply our analysis approach from Section \ref{sec:method}. In our work, we aim to answer the following questions:

\begin{enumerate}
    \item Does the training stage have an impact on the performance of common Laplace approximation and ICLA?
    \item What is the connection between the structure of the Fisher matrix (i.e. model curvature) and the data it is used to model?
    \item Can we explain the outstanding results of ICLA by linking them to the separability of classes in the embedding space?
    \item How do the smoother solutions (with lowered long-tailness), made possible by applying the Fisher penalty, change the comparison between LLLA and ICLA?
\end{enumerate}


\subsection{Training Stage Independence}\label{sec:training_stage}

At first thought, it might seem like the performance gap between LLLA and ICLA might be connected to the training stage, since curvature continuously changes throughout the training process. However, our empirical observations suggest that the phenomena is not linked to the training stage of a model. In Figure \ref{epochwise}, we build approximations for different training epochs and show that even at the start of training, there is a positive performance gap that persists between the standard last-layer Laplace approximation and ICLA.

\subsection{Fisher Eigenvalues}\label{sec:fisher_eig}

Firstly, we examine the eigenvalues $\lambda_i$ of empirical Fisher used in LLLA. We observe that almost all eigenvalues are small, and only a tiny part of them exhibits extremely large values, demonstrating clear long-tailness. We show the spectral distributions in Figure \ref{spectral}. Our observation perfectly aligns with the past research on analyzing the curvature of neural networks \cite{karakida2019universal, sagun2017empirical, lecun2002efficient}. This is a crucial property to consider, as matrices with long-tailed eigenvalues are used to model data covariance.

\subsection{Feature Embeddings}\label{sec:embeddings}

\begin{figure}[t]
\centering
\includegraphics[width=0.8\columnwidth]{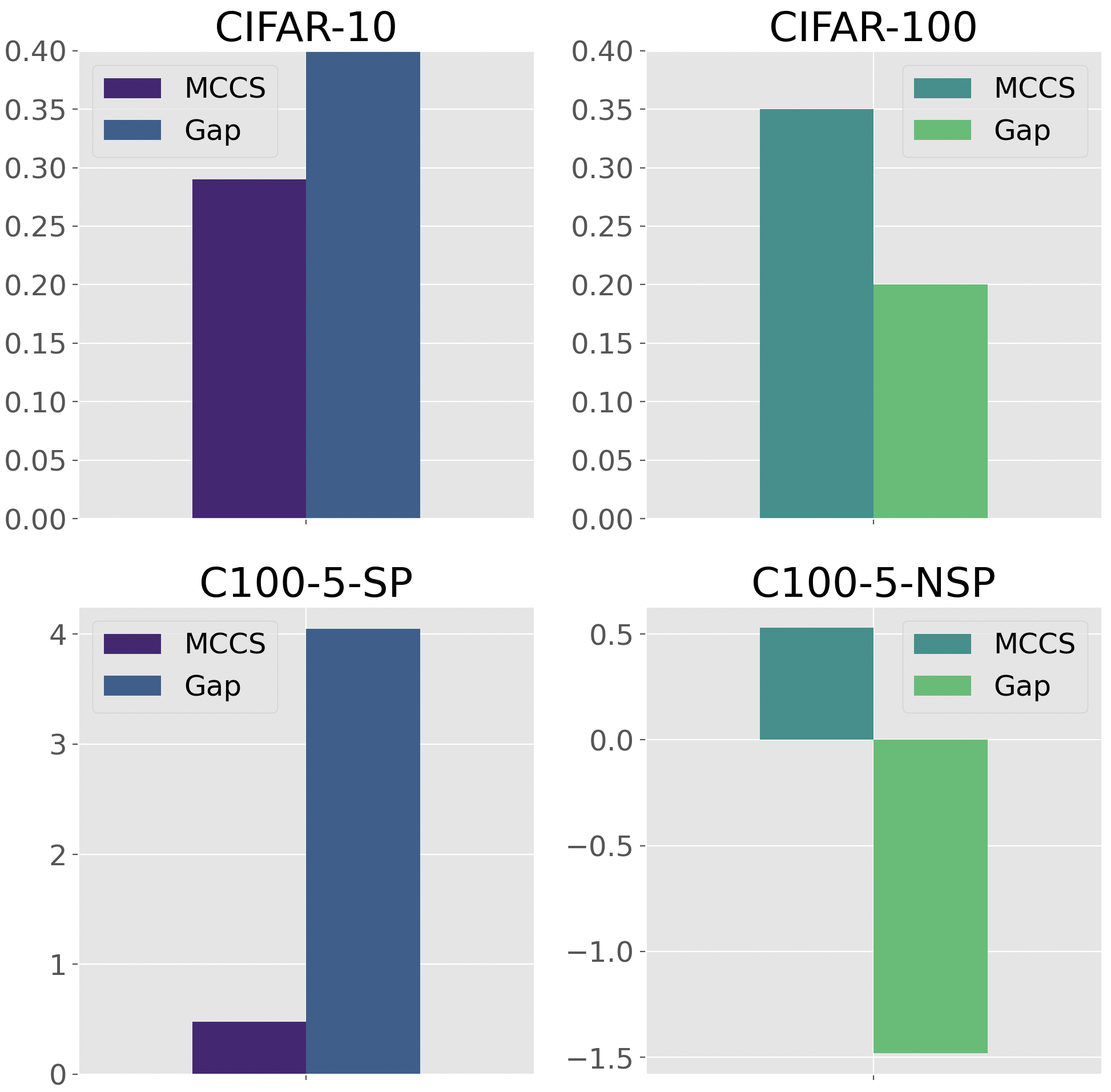} 
\caption{MCCS and performance gap comparison for CIFAR-10, CIFAR-100, C100-5-SP and C100-5-NSP. It can be seen that CIFAR-10 and C100-5-SP have a lower MCCS and larger gap, while CIFAR-100 and C100-5-NSP have a higher MCCS and lower gap. This points to a correlation between feature separability and Laplace approximation performance. See Section \ref{sec:embeddings} for details.}
\label{img:mccs_1}
\end{figure}

As one might recall from Equation \ref{eq:output_distribution}, the curvature information directly influences resulting uncertainty through covariance. This can be seen since $\Sigma^{-1}$ holds the \textit{congruence relation} to the resulting covariance of output distribution, i.e., the covariance of output distribution and Fisher represent the same linear transformation, but in different bases.

\begin{figure}[t]
\centering
\includegraphics[width=0.8\columnwidth]{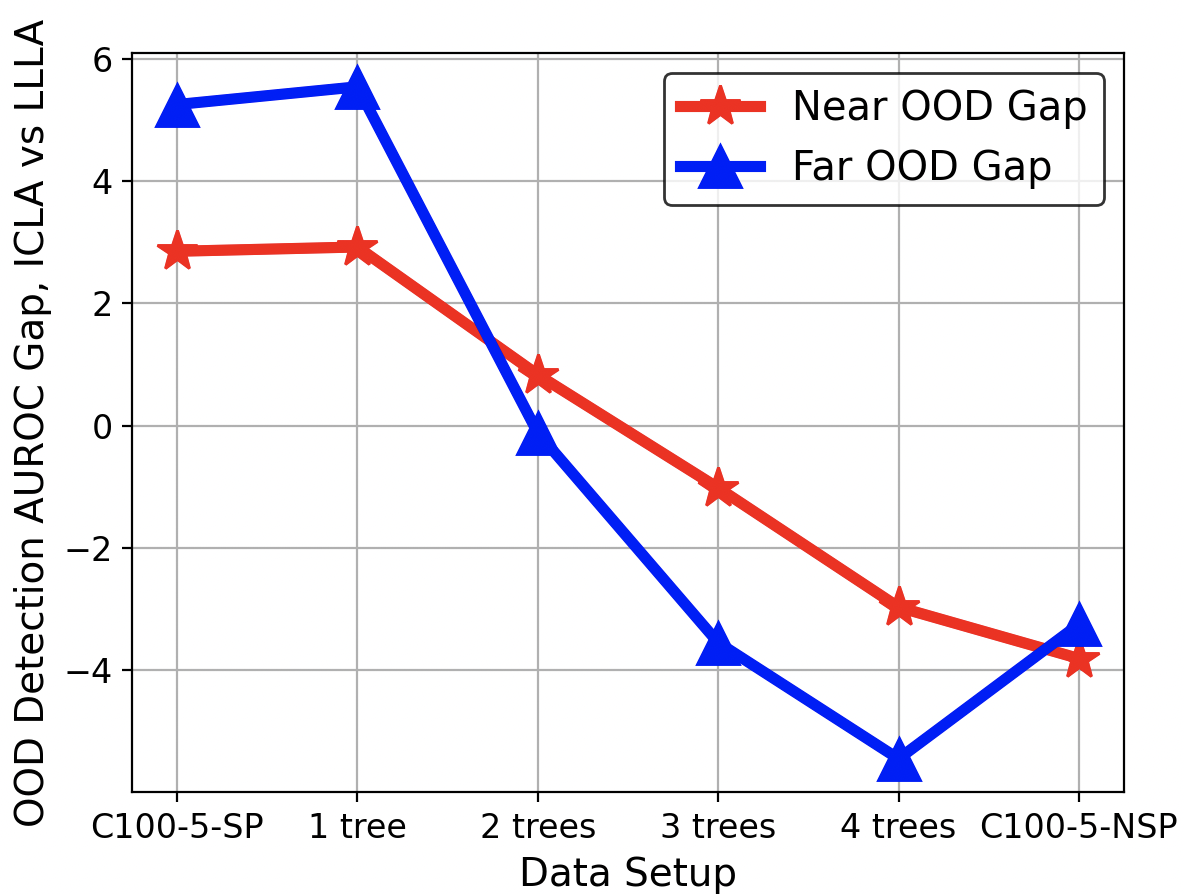} 
\caption{Data separability experiment. Gradually moving from a separable (C100-5-SP) to less separable dataset (C100-5-NSP), we observe how standard LLLA begins to estimate uncertainty better than its alternative with identity curvature. This can be seen as a sign of high data separability and curvature information being linked. See Section \ref{sec:penalty} for details.}
\label{img:trees}
\end{figure}

Naturally, this raises the following question: \textit{How does long-tailed curvature align with class covariance?} To answer this, we select 4 datasets with different inter-class separability: \textbf{CIFAR-10}, \textbf{CIFAR-100}, \textbf{C100-5-SP}. We use 5 classes from different super classes of CIFAR-100, namely “flowers”, “food containers”, “fruit and vegetables”, “household electrical devices” and “insects” (more separable case). From \textbf{C100-5-NSP}, we use 5 classes from the “trees” super class (less separable case).

\begin{figure*}[ht!]
\centering
\includegraphics[width=0.9\textwidth]{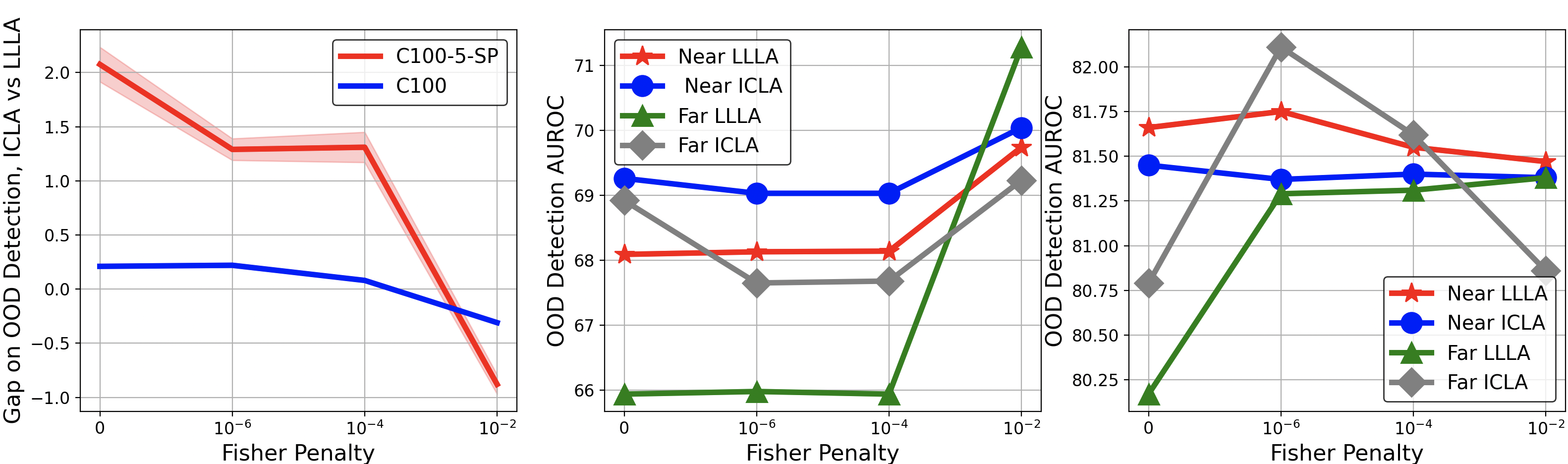} 
\caption{Effects of Fisher penalty. (\textit{Left}) OOD detection performance gap for ICLA and LLLA. (\textit{Middle}) OOD detection AUROC for the C100-5-SP dataset. (\textit{Right}) OOD detection AUROC for the CIFAR-100 dataset. As regularization increases, the gap vanishes. While the Fisher penalty improves both LLLA and ICLA, the effect on LLLA is more visible. See Section \ref{sec:penalty} for details.}
\label{img:fp_dynamics}
\end{figure*}

We use Equation \ref{eq:mccs} as a measure of similarity and separability between clusters of classes in the embedding space and compare it to the performance gap in OOD detection between LLLA and ICLA. The performance gap is calculated simply as the average of “Near OOD” and “Far OOD” performance differences between ICLA and LLLA.

Our observations are shown in Figure \ref{img:mccs_1}. From these results, we conclude that there is a trend of class separability being connected to how informative curvature is about data covariance. To visually examine separability, we demonstrate feature embeddings in Appendix \ref{app:embeddings}.

We conduct another experiment, which clearly shows the trend between LLLA performance and data separability. Starting from C100-5-SP, we train 5 models, and each of them receives one more class from the “trees” super-class than the previous one, replacing one class from C100-5-SP. Here, the first model is trained on C100-5-SP and the last model is trained on C100-5-NSP. This experimental setup allows us to show the trend between changing separability and performance gaps for ICLA and LLLA. Refer to Figure \ref{img:trees} for details. As one can see from our results, gradually moving from the most to the least separable case makes LLLA perform better than standard ICLA. The results clearly show that the efficacy of identity curvature can be explained by data separability.

\subsection{Smooth Curvature Gap Reduction}\label{sec:penalty}

Previously, we demonstrated the connection between data separability, curvature, and uncertainty estimation. Let us take a final step and connect the line between the performance of LLLA and smoothness.

\paragraph{Fisher Trace Penalty.} As was shown before, Fisher
with long-tailed spectral distribution is not informative for modeling class covariance, which leads to poor LLLA performance. Thus, for setups with highly separable embeddings, we would prefer the curvature to be smoother, i.e. lower Fisher long-tailness.

\begin{figure}[t]
\centering
\includegraphics[width=0.9\columnwidth]{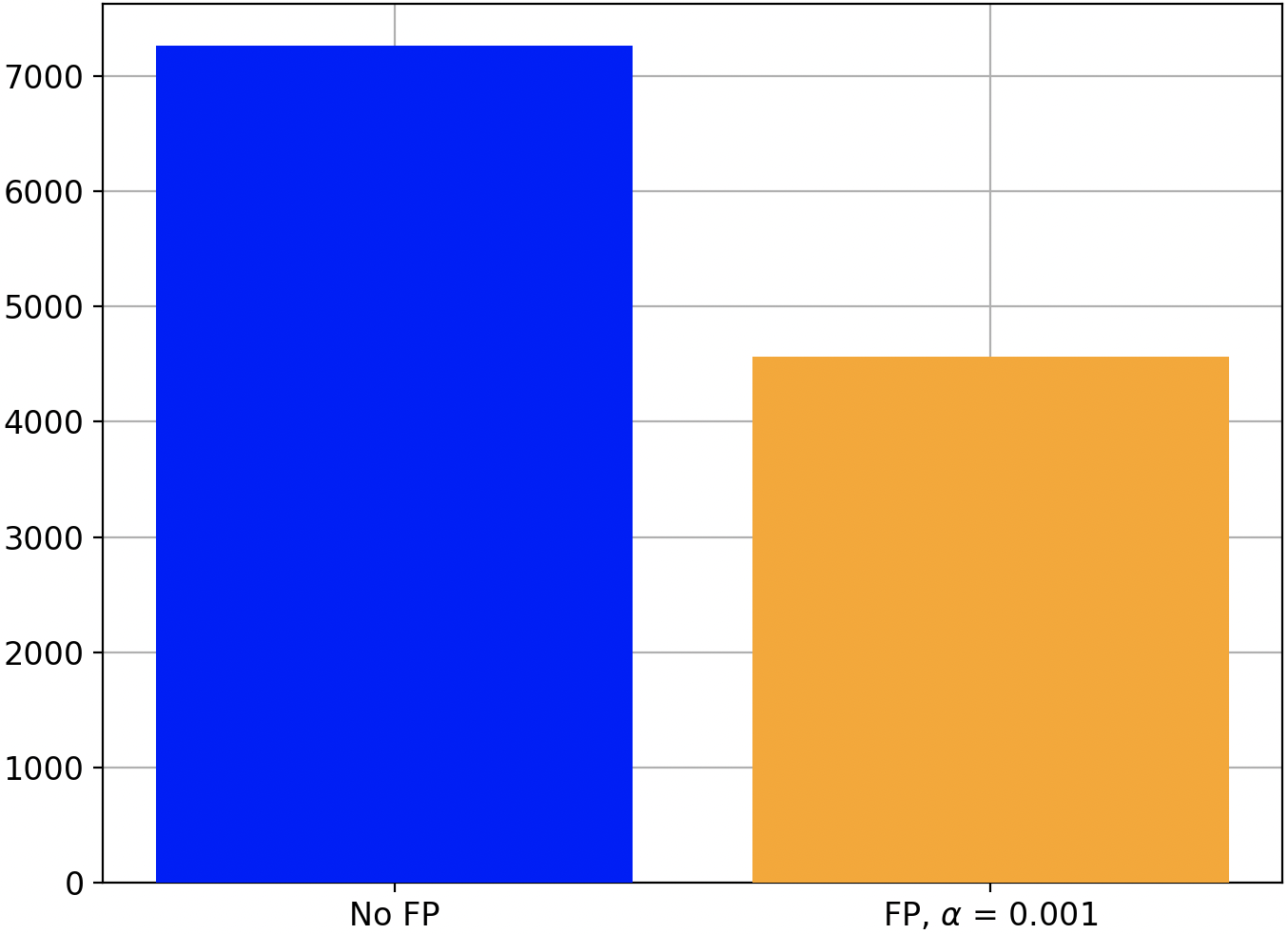} 
\caption{Mean eigenvalues of Fisher for a model trained on CIFAR-10 without the Fisher penalty and a model trained with the Fisher penalty, $\alpha = 0.001$. The model trained with the Fisher penalty achieves a flatter solution. See Section \ref{sec:penalty} for details.}
\label{img:eig_mean}
\end{figure}

Given the CIFAR-100 and C100-5-SP datasets with smaller separability from Section \ref{sec:embeddings}, we apply the Fisher penalty (Equation \ref{eq:fp}), progressively increasing $\alpha$, which changes the performance of LLLA and ICLA. Figure \ref{img:fp_dynamics} shows the dynamics of the vanishing gap between LLLA and ICLA. This highlights the connection between curvature and data separability. In Figure \ref{img:eig_mean}, we show the mean eigenvalues of Fisher for an unregularized model and a model trained with Fisher penalty. As one can see, it indeed reduces the magnitude of eigenvalues, leading to a flatter curvature.

\paragraph{Sharpness-Aware Minimization.}
In the context of our study, it can applied for the same reason as the Fisher Penalty, as we hypothesize that achieving flatter solutions should reduce the OOD detection performance gap. We employ training with ASAM for both LLLA and ICLA and compare its performance with Fisher Penalty. The results are presented in Table \ref{tab:sam}. Our observation show that applying ASAM also reduces the performance gap. 

\begin{table}
    \centering
    \caption{OOD detection results with Adaptive Sharpness-Aware Minimization (ASAM) for the CIFAR-10 dataset.}
    \begin{tabular}{c|c|c} \hline 
         &  \textbf{Near OOD}& \textbf{Far OOD}\\ \hline 
         LLLA+FP&  89.37& 91.91\\ 
         LLLA+ASAM&  91.92& 93.88\\ \hline
 Gap with FP& 0.62&0.60\\
 Gap with ASAM& 0.40&0.10\\\hline
    \end{tabular}
    \label{tab:sam}
\end{table}

The results for the Fisher Penalty and ASAM confirm our intuition regarding the performance gap between LLLA and ICLA. These experiments show that the observed gap is a curvature-related issue. Our analysis suggests that, when the data is highly separable, the model has high curvature (long-tailed Fisher), and achieving a flat curvature landscape improves LLLA performance and reduces the performance gap between LLLA and ICLA.

\section{Conclusion}

In this work, we demonstrated the effectiveness of Identity Curvature Laplace Approximation (ICLA) in improving Laplace approximation for out-of-distribution (OOD) detection. Our empirical results indicate that ICLA significantly enhances OOD detection performance on the CIFAR-10, CIFAR-100, and ImageNet-200 datasets, while maintaining adequate calibration scores. We established a link between ICLA performance and class separability in the embedding space. Our findings suggest that improving the OOD detection capability of Laplace approximation can have significant implications for developing safe and deployable AI systems.

\paragraph{Limitations.} While our study suggests that using empirical Fisher (EF) Hessian approximation to study the curvature is inconsequential for the obtained results, more work can be done to confidently confirm it.

{\small
\bibliographystyle{ieee_fullname}
\bibliography{egbib}
}

\clearpage

\thispagestyle{empty}
\appendix

\section{Marginal Likelihood Algorithm}\label{app:algs}

We present a detailed definition of the \verb|marginal_likelihood| function in Algorithm \ref{alg:marglik}.

\begin{algorithm}
\caption{Marginal Likelihood}\label{alg:marglik}
\begin{algorithmic}[1]

\Require{Dataset $\mathcal{D}=\{x_i, y_i\}^N_{i=1}$, neural network $f$, Hessian $\mathcal{H}$, learning rate $\alpha$, number of epochs $T$.}
\Ensure{Prior precision $\lambda$ for Equation 
\ref{eq:icla}.}
\State Initialize $\lambda$
\For{$t=1,\dots,T$}
    \State $h_t \gets \nabla_{\lambda^2}\operatorname{log}(p(\mathcal{D}|f, \lambda^2))$
    \State $\lambda^2_t \gets \lambda^2_{t-1} + \alpha h_t$
\EndFor
\State \Return $\lambda$

\end{algorithmic}
\end{algorithm}

\section{Computational performance comparison}\label{app:compute}

\begin{table}
\centering
\caption{Computational costs of ICLA and LLLA. The results for initialization are reported in seconds per initialization and seconds per 1000 batches of size 64 for the inference stage.}
\begin{tabular}{@{}cccc@{}}
\toprule
&\multicolumn{3}{c}{Inference}\\
        & CIFAR-10    & CIFAR-100   & ImageNet\\ \midrule
EF      & 4.69        & 10.81       & 10.68\\
GGN     & 4.77        & 10.51       & 10.56\\
KFAC    & 5.11        & 30.22       & 29.31\\
ICLA    & 4.65        & 10.70       & 10.41\\ \midrule
&\multicolumn{3}{c}{Initialization}\\
        & CIFAR-10    & CIFAR-100   & ImageNet\\ \midrule
EF      & 19.71       & 20.68       & 21.38\\
GGN     & 19.28       & 32.6        & 37.07\\
KFAC    & 19.38       & 21.44       & 20.46\\
ICLA    & 0.02        & 0.02        & 0.05\\ \bottomrule
\end{tabular}
\label{tab:compute}
\end{table}

We report computational performance comparison in Table \ref{tab:compute}. All the comparisons are conducted on a single A100 80GB GPU. It can be seen that our ICLA implementation doesn't impose any additional computational overhead on inference and takes near zero time for initialization (Algorithm \ref{lst:icla}).

\section{Discussion on the connection of ICLA and NECO}
NEural Collapse-based Out-of-distribution (NECO \cite{ammar2023neco}) has a theoretical framework based on class separability and analyzing eigenvalues of covariance matrix in embedding space. Although some of these aspects might seem related, our analysis method focuses on connecting class separability with the Fisher matrix structure and model curvature, which allows us to put our approach into the field of Laplace approximation and Bayesian methods.

\section{Calibration Metrics}\label{app:calibration}

In this section, we list performance measurements for calibration. 
Firstly, let $B_t$ be a batch of samples, whose confidences lie in the interval $( \frac{t-1}{T}, \frac{t}{T} ]$, where $T$ is the number of bins we split the prediction by and $m$ is the bin index. We define accuracy and confidence as 

\begin{equation}
    \text{acc}(B_t) = \frac{1}{|B_t|}\sum_{i \in B_t} \mathbb{I}(f(x_i) = y_i). 
\end{equation}
\begin{equation}
    \text{conf}(B_t) = \frac{1}{|B_t|}\sum_{i \in B_t}p_i.
\end{equation}

\subsection{Expected Calibration Error}

The expected calibration error (ECE) can be estimated as 

\begin{equation}
    \text{ECE}(B_t) = \sum_{t=1}^T \frac{|B_t|}{N} |\text{acc}(B_t) - \text{conf}(B_t)|
\end{equation}

where $N$ is the number of samples.

\subsection{Negative Log-Likelihood}

Negative log-likelihood typically coincides with cross-entropy and is computed as

\begin{equation}
    \text{NLL} = -\sum_{i=1}^{N}\operatorname{log}(f(x_i).
\end{equation}

\subsection{Brier Score}

Brier score is another calibration measure and is expressed as

\begin{equation}
    \text{Brier} = \frac{1}{N}\sum_{i=1}^{N}(f(x_i) - y_i)^2.
\end{equation}

\section{Details on Hessian Approximations}\label{app:hessian}

In this section, we elaborate on Hessian approximations and their formulations.

\subsection{Generalized Gauss-Newton Matrix (GGN)}

\begin{equation}
    G \triangleq \sum^N_{i=1} J(x_i) (\nabla^2_{\theta} \operatorname{log} p(y_i | f_{\theta}|_{\theta_{\text{MAP}}}) J(x_i)^T,
\end{equation}

where $J(x) \triangleq \nabla^2_{\theta}\operatorname{log}(p(\mathcal{D}|\theta))|_{\theta_{\text{MAP}}}$ is the Jacobian matrix.



\subsection{K-FAC}

In some setups, Fisher might still require solid computational resources. Another popular factorization is the Kronecker-factored approximate curvature (K-FAC). It factorizes layer-wise Fisher as a Kronecker product of smaller matrices under the assumption of independence of layer-wise weights.

Given a layer with $N$ hidden units, denote $h_n$ as the n-th hidden vector and $g_n$ as the log-likelihood gradient w.r.t. $h_n$. Then Fisher can be approximated as

\begin{equation}
    F_n \approx \mathbb{E} (h_{n-1}h_{n-1}^T) \otimes \mathbb{E} (g_n g_n^T).
\end{equation}



\section{Embeddings Visualization}\label{app:embeddings}

\begin{figure*}[ht!]
\centering
\includegraphics[width=0.9\textwidth]{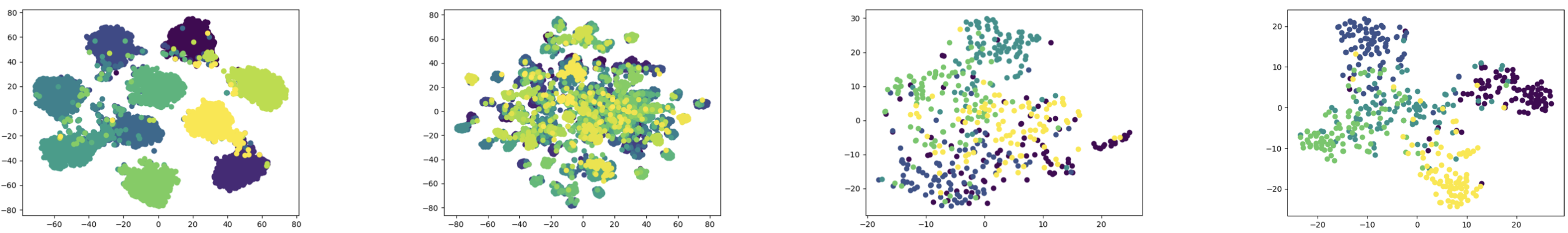} 
\caption{Visualizations of feature embeddings. (\textit{First}) CIFAR-10. (\textit{Second}) CIFAR-100. (\textit{Third}) C100-5-NSP. (\textit{Fourth}) C100-5-SP. CIFAR-10 and C100-5-SP present more separability, as the clusters of classes overlap less. ICLA performs better on more separable cases, showing the connection between curvature and data separability. See Section \ref{sec:embeddings} for details.}
\label{fig:emb_vis}
\end{figure*}

In this section, we visualize the embeddings for CIFAR-10, CIFAR-100, C100-5-SP and C100-5-NSP in Figure \ref{fig:emb_vis}.

\section{Additional Calibration Details}

Here, we provide a precise comparison between LLLA variations and ICLA for calibration in Table \ref{tab:fullcalib}.

\section{Prior Precision Values}

We report the obtained prior precision $\lambda$ values in Section \ref{sec:ood}: 2.76 for CIFAR-10 and 3.06 for CIFAR-100.

\section{Prior Precision Impact on OOD Detection}

We demonstrate the relation between the value of prior precision $\lambda$ and OOD detection AUROC in Table \ref{tab:lambda}. As can be seen, prior precision values affect the OOD detection performance. It makes prior precision optimization sensible in our algorithm.

\begin{table}
    \centering
    \caption{OOD detection AUROC depending on prior precision value $\lambda$ for CIFAR-10 dataset.}
    \begin{tabular}{c|c|c}
         \hline
         \textbf{$\lambda$}&  \textbf{Near OOD}& \textbf{Far OOD}\\\hline
         1&  90.31& 92.20\\
         3&  89.89& 91.76\\
         5&  89.06& 91.08\\
         7&  88.84& 90.86\\
         \hline
    \end{tabular}
    \label{tab:lambda}
\end{table}

\begin{table*}[ht!]
\renewcommand{\arraystretch}{1.2} 
\setlength{\tabcolsep}{5pt} 
\centering
\caption{Precise calibration results between LLLAs and ICLA.}
\vskip 0.15in
\begin{tabular}{@{}c|c|cccc@{}}
\toprule
Dataset & Metric & LLLA (GGN) & LLLA (EF) & LLLA (K-FAC) & ICLA \\ \hline
\multirow{3}{*}{CIFAR-10} & ECE & 1.23 ± 0.21 & 10.80 ± 7.10 & 2.75 ± 0.29 & 15.15 ± 4.53 \\
 & NLL & 1.53 ± 0.01 & 1.62 ± 0.05 & 1.51 ± 0.03 & 1.59 ± 0.04 \\
 & Brier & 0.07 ± 0.01 & 0.09 ± 0.01 & 0.08 ± 0.01 & 0.10 ± 0.01 \\ \hline
\multirow{3}{*}{CIFAR-100} & ECE & 19.11 ± 1.69 & 61.84 ± 4.08 & 6.47 ± 0.86 & 65.11 ± 0.54 \\
 & NLL & 4.08 ± 0.01 & 4.47 ± 0.03 & 3.89 ± 0.02 & 4.47 ± 0.03 \\
 & Brier & 0.37 ± 0.01 & 0.75 ± 0.05 & 0.32 ± 0.01 & 0.80 ± 0.01 \\ \hline
\multirow{3}{*}{ImageNet-200} & ECE & 3.54 ± 0.02 & 4.78 ± 0.24 & 1.77 ± 0.18 & 4.77 ± 0.26 \\
 & NLL & 3.53 ± 0.01 & 3.91 ± 0.01 & 3.11 ± 0.01 & 3.94 ± 0.01 \\
 & Brier & 0.21 ± 0.01 & 0.22 ± 0.01 & 0.20 ± 0.01 & 0.22 ± 0.01 \\ \bottomrule
\end{tabular}
\vskip -0.1in
\label{tab:fullcalib}
\end{table*}

\end{document}